\documentclass[letterpaper, 10 pt, conference]{ieeeconf}  

\IEEEoverridecommandlockouts                              

\usepackage{graphicx}
\usepackage{subfig}
\graphicspath{{Figures}}
\DeclareGraphicsExtensions{.pdf,.png}

\usepackage{multirow}
\usepackage{bigstrut}
\usepackage{romannum}

\usepackage{amsmath}
\usepackage{amssymb}
\usepackage{bbm}
\usepackage{algorithm}
\usepackage[noend]{algorithmic}
\usepackage{makecell}
\usepackage{wrapfig,lipsum,booktabs}
\setcellgapes{1pt}
\usepackage[font=footnotesize]{caption}
\usepackage{float}
\usepackage{stfloats}
\let\labelindent\relax
\usepackage{enumitem}
\usepackage[nodisplayskipstretch]{setspace}
\usepackage{hyperref}

\newcommand{\argmin}{\arg\!\min}
\newcommand{\subshrink}{\vspace{-5pt}}
\newcommand{\shrink}{\vspace{-10pt}}

\newcommand{\shrinkbottom}{\vspace{-7pt}}

\newcommand{\eref}[1]{Eqn.~(\ref{#1})}  
\newcommand{\sref}[1]{Sec.~\ref{#1}}    
\newcommand{\figref}[1]{Fig.~\ref{#1}}  
\newcommand{\prg}[1]{\noindent\textbf{#1. }} 
\newcommand{\mx}[1]{\textcolor{black}{#1}}

\newcommand{\btheta}{{\pmb{\theta}}}
\newcommand{\bs}{{\pmb{s}}}

\usepackage[colorinlistoftodos,backgroundcolor=yellow,bordercolor=yellow,linecolor=yellow,textcolor=gray,textsize=scriptsize]{todonotes}

\definecolor{teal}{rgb}{0.26,0.52,0.56}
\definecolor{orange}{rgb}{0.89,0.55,0.06}

\title{\LARGE \bf Learning Tool Morphology for Contact-Rich Manipulation Tasks\\ with Differentiable Simulation}
\author{
  Mengxi Li,
   Rika Antonova,
   Dorsa Sadigh and 
   Jeannette Bohg
\thanks{The authors are with the Department of Computer Science at Stanford University (contact: mengxili@stanford.edu, rika.antonova@stanford.edu, dorsa@cs.stanford.edu, bohg@stanford.edu). This project was supported in part by FANUC. Rika Antonova is supported by the National Science Foundation grant No.2030859 to the Computing Research Association for the CIFellows Project. This work is to appear in the International Conference on Robotics and Automation (ICRA) 2023, copyright IEEE.}}

\definecolor{mydarkblue}{rgb}{0,0.08,0.45}
\hypersetup{ %
    pdftitle={Learning Tool Morphology for Contact-Rich Manipulation Tasks with Differentiable Simulation},
    pdfauthor={Mengxi Li, Rika Antonova, Dorsa Sadigh, Jeannette Bohg},
    pdfsubject={Robotics},
    pdfkeywords={Differentiable simulation, Morphology optimization, Deformable objects},
    pdfborder=0 0 0,
    pdfpagemode=UseNone,
    colorlinks=true,
    linkcolor=mydarkblue,
    citecolor=mydarkblue,
    filecolor=mydarkblue,
    urlcolor=mydarkblue,
    pdfview=FitH
}

\begin{document}

\maketitle
\thispagestyle{empty}
\pagestyle{empty}

\begin{abstract}
    When humans perform contact-rich manipulation tasks, customized tools are often necessary to simplify the task.  For instance, we use various utensils for handling food, such as knives, forks and spoons. Similarly, robots may benefit from specialized tools that enable them to more easily complete a variety of tasks. We present an end-to-end framework to automatically learn tool morphology for contact-rich manipulation tasks by leveraging differentiable physics simulators. Previous work relied on manually constructed priors requiring detailed specification of a 3D object model, grasp pose and task description to facilitate the search or optimization process. Our approach only requires defining the objective with respect to task performance and enables learning a robust morphology through randomizing variations of the task. We make this optimization tractable by casting it as a continual learning problem. We demonstrate the effectiveness of our method for designing new tools in several scenarios, such as winding ropes, flipping a box and pushing peas onto a scoop in simulation. Additionally, experiments with real robots show that the tool shapes discovered by our method help them succeed in these scenarios.
\end{abstract}
\section{Introduction}

Humans are distinct from other species in that tool use is a defining and universal characteristic~\cite{gibson1994tools}. This suggests that in the pursuit of equipping robots with human-like dexterity, tools may play an important role. Robots are already using various tools in a range of contact-rich manipulation tasks. For example, to make knots, robots can use a tri-needle to maintain the loop \cite{saha2007manipulation}. For cooking, robots use spatulas to flip pancakes \cite{tsuji2015dynamic} and skewers to pick food for assistive feeding \cite{sundaresan2022learning, grannen2022learning}. While tools greatly influence how robots interact with the environment in these contact-rich tasks, most works focus on learning how to use existing tools. Little attention is paid to optimal tool design. Rather than forcing robots to use pre-defined tools, we aim to intelligently adapt the tools to the tasks, thus helping robots become more effective.

In this work, we aim to develop a general framework for learning robust tool morphology for contact-rich tasks.
Relevant to our setting, works on aerodynamic design \cite{lyu2014benchmarking} and vehicle component design also tackle the problem of finding an optimal shape. However, these prior works do not focus on tool design for contact-rich tasks and thus do not require complex contact modeling that is necessary in our scenarios. Another relevant line of research investigates robot gripper design. Some of these works aim to discover gripper designs for grasping a wide range of objects \cite{zhang2016dorapicker} or executing re-orientation primitives~\cite{rojas2016gr2}. However, they do not provide a way to optimize the gripper shape for more complex tasks. 
Recent work~\cite{xu2021end} proposes a new approach to optimize morphology for a given task objective.
However, as we will show in our experiments, it does not necessarily generalize to task variations, such as different initial object poses. 
We aim to automatically design tools for a given task objective, such that these tools are robust to task variations.

\begin{figure}[]
	\begin{center}
		\includegraphics[width=\columnwidth]{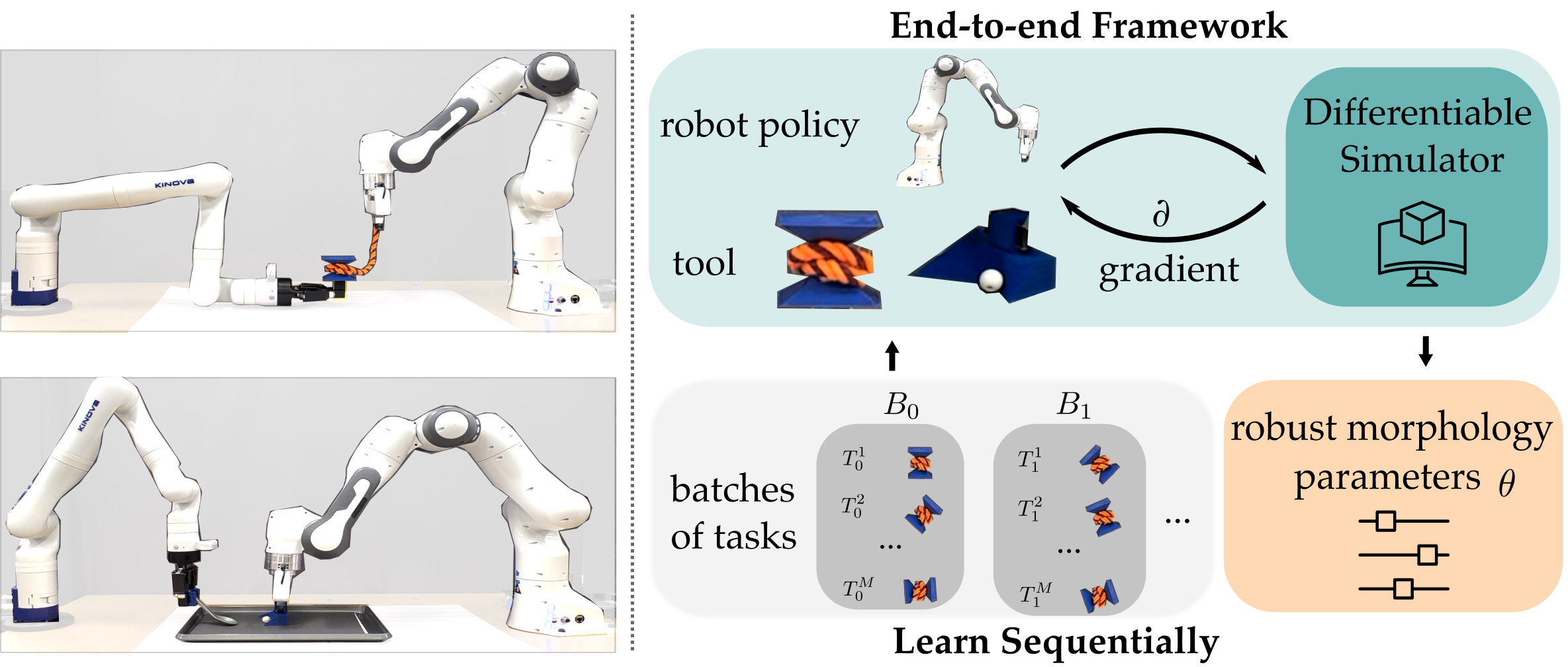}
		\caption{We build an end-to-end framework for learning tool morphologies suitable for contact-rich tasks. Our goal is to learn a tool morphology for a given scenario that is robust to task variations. We achieve this with a method based on continual learning that trains on a sequence of task variations.}
		\label{fig:frontfig}
		\shrinkbottom
	\end{center}
\vspace{-15pt}
\end{figure}

To obtain an optimal shape that minimizes a task-relevant objective, previous work on gripper design employed heuristics for guiding their search \cite{gorce1996design}, or needed Monte Carlo estimation of the gradients to attempt gradient-based optimization~\cite{saravanan2009evolutionary}. When a robot uses a tool in a contact-rich manipulation task and knows the corresponding dynamics model, it is possible to directly get the exact gradients by differentiation; this enables better numerical stability and faster convergence~\cite{de2018end}. To this end, we leverage recent advances in differentiable physical simulators \cite{degrave2019differentiable, hu2020difftaichi, werling2021fast, xu2021end} and build an end-to-end framework for learning tool morphology suitable for contact-rich tasks. 

To design an effective tool for a specific scenario while maintaining the ability to handle task variations, we optimize tool morphology over a distribution of task variations. This presents two challenges. First, the training process for learning to handle the entire distribution can be computationally expensive. Second, due to the complex dynamics of contact-rich tasks, the underlying optimization landscape is highly non-linear and rugged, which makes it more difficult for optimization to converge to a good solution~\cite{suh2022differentiable,antonova2022rethinking}. To tackle these challenges, we propose an approach based on continual learning that samples task variations and conducts optimization in a sequential manner. Our insight is that we can re-interpret continual learning as a robust optimization framework for problems with challenging loss landscapes. \mx{Compared to prior work, we broaden morphology optimization to tool design, thereby opening up a path for tackling a wider array of tasks and objects, including deformables.}
Furthermore, we show that the tools obtained with our method are effective for completing the given tasks in reality.

\section{Related Work}

\prg{Morphology Optimization in Manipulation} In this work, we explore the promise of fully differentiable end-to-end optimization of tool morphology with the help of differentiable physics. 
\textit{\mbox{Gripper design}} is one related problem. Some classic works provide guidelines for manually designing grippers guided by practical insights~\cite{causey1998gripper}. Recent works aim to learn an optimal design within a given design space. Evolutionary strategies are employed in \cite{meixner2019automated} to optimize both the robot morphology and the controller. In~\cite{wolniakowski2015task}, the authors optimize gripper quality using a set of manually designed metrics. Another popular paradigm for gripper customization is imprint-based methods \cite{Schwartz2017DesigningFI, velasco1998computer}. However, these methods only produce customized finger geometry, where an object 3D model, grasp pose, and task description are specified by the user. A survey in~\cite{honarpardaz2017finger} reviews other examples of automated finger design. Further recent examples include a gradient-free method~\cite{pan2021emergent}, and a gradient-based shape generation method with non-differentiable simulation for training~\cite{ha2020fit2form}. Most closely related to our work is DiffHand~\cite{xu2021end}, where a differentiable simulator is developed to enable co-design of robot morphology and control by optimizing task-specific objectives. However, this work only considers specific object initial states, and thus does not learn a morphology that would generalize over various initial object states. 

\textit{Tool design} is related to gripper design but unlike grippers and fingers, tools are typically not rigidly attached to robots. This opens future possibilities to study a) re-grasping of tools to improve versatility and dexterity of manipulation robots, and b) the use of multiple tools at the same time (e.g. for dual-arm manipulation). While we do not focus on these aspects in our current work, our formulation facilitates exploring them in the future.
Works that consider selecting tools and optimizing policies for tool use are common in robotics literature, e.g.~\cite{shao2020learning, toussaint2018differentiable}, where \cite{toussaint2018differentiable} employs a differentiable simulator. However, the vast majority of these works do not optimize tool shape. Related work for tool \textit{morphology optimization} includes MacGyvering~\cite{nair2019tool}. However, it assumes access to a `reference tool', and aims to construct a tool from a set of available parts. Instead, we consider the problem of evolving tool morphology from an initial shape without assuming prior knowledge about the optimal tool shape.  Furthermore, we use end-to-end differentiable simulation, and leverage differentiability at all levels of the optimization pipeline.

\prg{Differentiable Simulation} 
Instead of using heuristics or search algorithms for optimizing morphology, we leverage differentiable simulation for directly obtaining analytical gradients with respect to the final task objective. Several differentiable simulation frameworks have been developed recently, e.g. \cite{degrave2019differentiable, hu2020difftaichi, werling2021fast, xu2021end}. In this work, we adopt the differentiable simulator and morphology representation from DiffHand \cite{xu2021end}. However, our framework is agnostic to the choice of differentiable simulator. Differentiable simulation already enabled some impressive results: e.g. system identification (real-to-sim) and control optimization for cutting~\cite{heiden2021disect}; solving a dynamic ball-in-cup task in 4 minutes on a real robot~\cite{lutter2021differentiable}.
These works consider advanced phenomena, such as modeling deformation. However, they do not address the aspect that could be especially challenging for computing gradients -- making and breaking contacts in contact-rich tasks. Contact-rich scenarios bring a new level of complexity. They yield sharp changes in the loss landscapes and could make gradient-based optimization difficult~\cite{suh2022differentiable, antonova2022rethinking}. Our focus on tool morphology requires us to address this challenge because tools interact with objects in the scene. Hence, we propose a method that not only employs differentiability, but also leverages a continual learning formulation of the problem to tackle the optimization challenges that arise in contact-rich scenarios.

\prg{Continual Learning} Continual Learning considers the problem of learning to solve a sequence of tasks (or task variations), with the objective to perform well on the current task without forgetting what has been learned from previous tasks~\cite{delange2021continual, ha2020fit2form, lesort2020continual}. 
The tasks are usually provided as a stream, i.e. data from previous tasks is usually not retained due to memory limitations. There are different categories of continual learning methods. For example, replay methods~\cite{rebuffi2017icarl, lopez2017gradient} usually store a fixed number of samples in the replay buffer and use these to construct a distillation loss between previous and current model predictions. This distillation loss encourages the model not to forget what it previously learned. Parameter isolation methods~\cite{ mallya2018packnet, serra2018overcoming} divide model parameters into different subsets and fix the subset of parameters learned with previous tasks when learning a new task. In this way, they prevent the model from forgetting previous tasks and also improve training stability. We build upon these works and leverage the insight that continual learning could be re-interpreted as a framework for robust optimization suitable for problems with non-smooth dynamics that yield challenging loss landscapes, such as optimizing the morphology of manipulation tools. 

\section{End-to-end Framework with Differentiable Simulation}
\label{sec:pipeline}
In this section, we describe how we learn tool morphology in an end-to-end manner. In \sref{sec:task_formulation}, we first formalize the task where the tool will be used. With the deformation based morphology parameterization described in \sref{sec:cbd}, we obtain a low-dimensional design space. Finally, we show the end-to-end morphology parameter learning pipeline for a single task variation in \sref{sec:e2epipeline}.

\subshrink
\subsection{Problem Statement}
\label{sec:task_formulation}
We formulate the overall manipulation task as a discrete-time Markov Decision Process (MDP) with state space $\mathcal{S}$, action space $\mathcal{A}$, reward $r$, discount factor $\gamma$, and $\rho_{0}$ as the distribution of the initial state $\bs_0$: $\left(\mathcal{S}, \mathcal{A}, \mathcal{T_\btheta}, r, \gamma, \rho_{0}\right)$. We parameterize the transition function $\mathcal{T_\btheta}(s, a)$ by $\btheta$, which denotes a vector of tool morphology parameters as will be detailed in \sref{sec:cbd}. Different tool morphologies influence how the robot has to interact with an object to maximize reward. Imagine we are pushing peas onto a scoop with a tool that has a simple rectangular shape. It is likely that the peas will roll off to the side requiring the robot to re-orient the tool while pushing. However, if the tool is shaped to have a concave recess in the center, the peas would be less likely to roll off and escape while being pushed. This illustrates that different $\btheta$s yield different MDPs, since they change the transition function $\mathcal{T_\btheta}(s, a)$. Our goal is to learn the optimal morphology parameters $\btheta^*$ that maximize the task reward when the robot executes a control policy $\pi$. In principle, this framework is general enough to allow joint learning of morphologies and control policies. In practice, in this work we focus on learning tool morphology\footnote{In this work, we focus on morphology learning and regard policy learning as a separate line of work. However, our framework is compatible with policy learning as well. We show in the supplement video that existing methods, e.g. DiffHand~\cite{xu2021end} have significant difficulties with joint optimization of policy and morphology when presented with task variations. Hence, this is still an open problem for future work. \mx{Please see the Appendix and the \href{https://sites.google.com/stanford.edu/learning-tool-morphology}{video} for more details on the policy training. }}. To be consistent with notation in the literature on differentiable simulation, instead of reward maximization we describe the optimization problem as loss minimization.
\subshrink
\subsection{Morphology Parameterization using Cage-Based Deformation}
\label{sec:cbd}
\begin{figure}[t!]
    \begin{center}
   \vspace{5px}
   \includegraphics[width=0.6\columnwidth]{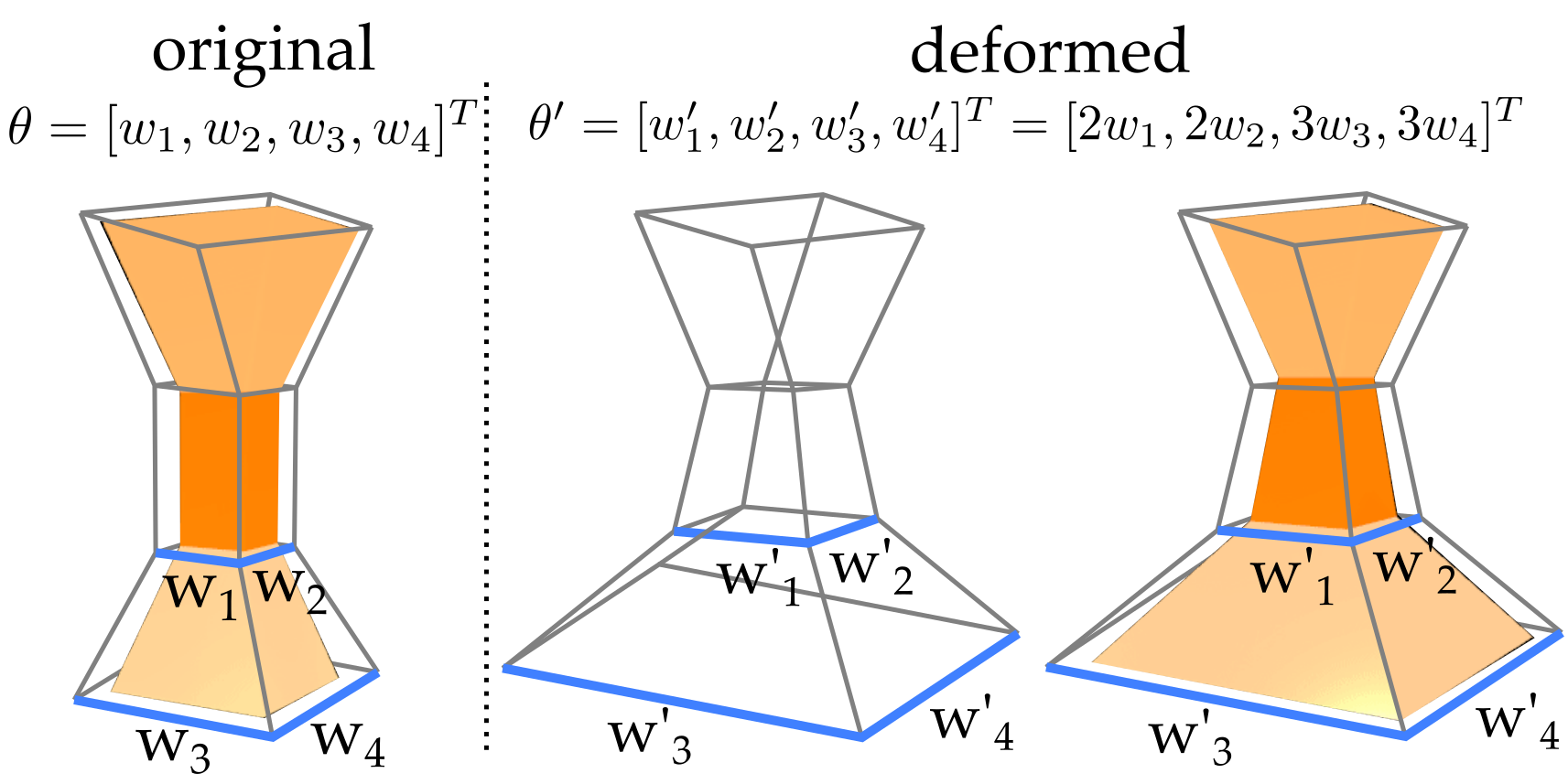}
		\caption{An example of morphology parameter vector $\btheta$ and shape deformation. Here, the morphology is parameterized by $\btheta = [\theta_1, \theta_2, \theta_3, \theta_4]^T$, where $\theta_1, \ldots, \theta_4$ represent the lengths of the four segments highlighted in blue. Upon updating morphology parameters $\btheta$ to $\btheta'$, we first get the deformed cage vertices, then get the full mesh, as described by~\eref{eq:cbd_mapping}. }
		\label{fig:cage}
		\shrinkbottom
		\shrink
	\end{center}
  \vspace{-1px}
\end{figure}
Cage-based deformation techniques are a common tool for deforming meshes in graphics applications \cite{nieto2013cage} and has previously been used to learn optimal hand morphology in a robotics context \cite{xu2021end}. We adopt the same morphology parametrization. Typically, a cage is a closed, low resolution mesh that envelopes the high-resolution mesh of the object we want to deform. Given an initial mesh $\mathcal{M}$ of the object and the cage $\mathcal{C}$ around it, we use $\mathbf{m_i} \in \mathcal{M}$ to denote the position of the $i^{th}$ mesh vertex, and $\mathbf{c_j} \in \mathcal{C}$ to denote the position of the $j^{th}$ vertex in the cage. Cage-based deformation establishes a linear mapping from the cage vertices $\mathbf{c_j} \in \mathcal{C}$ to each mesh point $\mathbf{m_i} \in \mathcal{M}$ by computing deformation weights $w_{ij}$:
\begin{equation} \label{eq:cbd_mapping}
    \mathbf{m_i} = \sum_{j, \mathbf{c_j}\in \mathcal{C}}w_{ij} \mathbf{c_j},  ~~~~~\sum_{j, \mathbf{c_j}\in \mathcal{C}}w_{ij} = 1, ~~~~~ \forall i, \mathbf{m_i} \in \mathcal{M}
\end{equation}
In this work, we adopt the Mean Value Coordinate method \cite{ju2005mean,xu2021end} for computing the deformation weights $w_{ij}$. Using these weights, we can manipulate the low-resolution cage vertices to deform the high-resolution mesh. Suppose we deform the cage $\mathcal{C}$ and obtain new cage vertices $\mathbf{c^\prime_j} \!\in\! \mathcal{C^\prime}$. Then, we can compute the deformed mesh vertices as $\mathbf{m^\prime_i} \in \mathcal{M^\prime}$, with $\mathbf{m^\prime_i} = \sum w_{ij} \mathbf{c^\prime_j},  \   \forall j, \mathbf{c^\prime_j}\in \mathcal{C^\prime}$.

To get an even more compact representation of the tool morphology, we further extract the high-level morphology attributes (e.g. tool segment length, height, width) and denote these morphology parameters as $\btheta \in \mathbb{R}^d$. \figref{fig:cage} shows a basic example. When we change the morphology parameters corresponding to segment width from $\btheta$ to $\btheta^\prime$, we first map $\btheta^\prime$ to a deformed cage $\mathcal{C}^\prime$, obtaining vertices $\mathbf{c^\prime_j}\in \mathcal{C}^\prime$. Then, we compute the deformed tool mesh $\mathbf{m^\prime_i}\in \mathcal{M}^\prime$ using \eref{eq:cbd_mapping}. The mappings from $\btheta'$ to cage $\mathcal{C}'$ and that from cage $\mathcal{C}'$ to object mesh $\mathcal{M}'$ are both linear, so the overall mapping from $\btheta'$ to mesh $\mathcal{M}'$ is linear as well.

\begin{figure}[t!]
  \begin{center}
  \vspace{5px}
  \includegraphics[width=0.9\columnwidth]{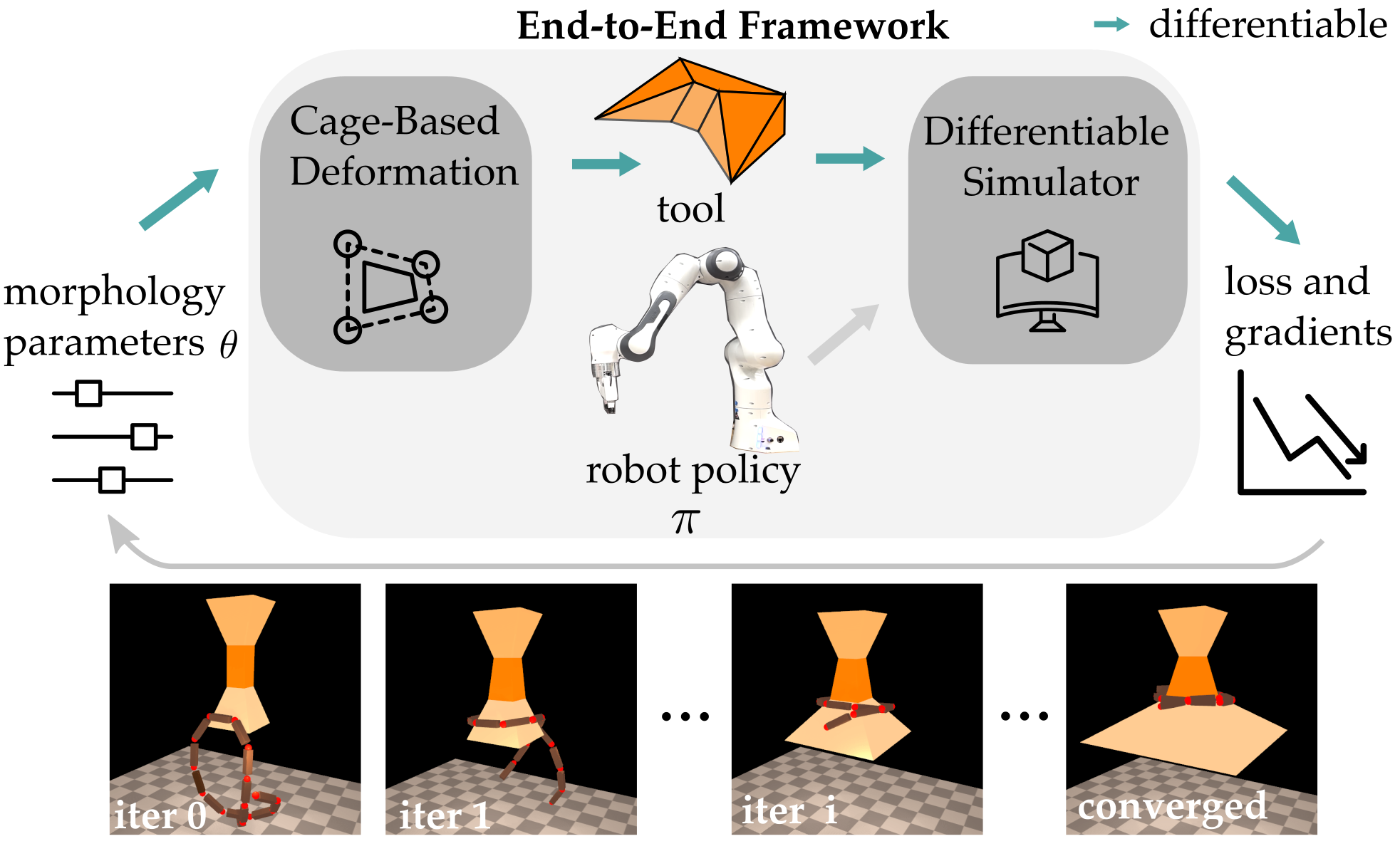}
		\caption{Visualization of our end-to-end pipeline for learning tool morphology. We first obtain the tool parametrized by morphology parameters $\btheta$ with cage-based deformation. Then we execute the policy $\pi$ with this tool in the simulator and output the loss. Since all of these operations are differentiable, we can get the analytical gradients to update $\btheta$. We also show the evolving shape for a winding tool. The bottom part gradually becomes larger, which prevents the rope from slipping off.}
		\label{fig:end2end}
		\shrinkbottom
		\shrink
	\end{center}
  \vspace{-1px}
\end{figure}

\subsection{End-to-end Pipeline Based on Differentiable Simulation}
\label{sec:e2epipeline}
With the morphology parameterized by Cage-Based Deformation, we can now build our end-to-end tool morphology learning framework with differentiable simulation. In this work, we adopt DiffHand \cite{xu2021end} as our simulator and build our pipeline upon it. However, our framework is agnostic to the choice of differentiable simulators and thus compatible with other differentiable simulators as well.

We visualize our end-to-end pipeline in \figref{fig:end2end}. With the cage-based deformation, we effectively parametrize the tool with morphology parameter $\btheta$. We then instantiate the task with the deformed tool, initialize it with state $\bs_0$, and execute the robot policy $\pi$ in the differentiable simulator. We then compute the gradients of the task-dependent loss $\mathcal{L}(\pi; \bs_0, \btheta)$ with respect to $\btheta$:
\begin{equation}
\frac{\partial \mathcal{L}(\pi; \bs_0, \btheta)}{\partial \btheta} = \textstyle\sum_{m_i \in \mathcal{M}, c_j \in \mathcal{C} }\frac{\partial \mathcal{L}(\pi; \bs_0, \btheta)}{\partial m_i} \cdot \frac{\partial m_i}{\partial c_j} \cdot \frac{\partial c_j}{\partial \btheta}.
\label{eq:gradient_bp}
\end{equation}
 $\frac{\partial \mathcal{L}(\pi; \bs_0, \btheta)}{\partial m_i}$ is provided by the differentiable simulator. $\frac{\partial m_i}{\partial c_j}$ and $\frac{\partial c_j}{\partial \btheta}$ are derived from the cage-based deformation in \sref{sec:cbd}.
We then take a gradient step on the tool morphology parameter $\btheta$ and repeat the process until convergence.

\vspace{10px}
\section{Continual Learning \\for Robust Tool Morphology}
\label{sec:continuallearning}
In this section, we describe how we learn a robust tool morphology using the end-to-end pipeline from \sref{sec:pipeline}. We first elaborate on the challenges of learning robust tool morphology for contact-rich scenarios from several perspectives in \sref{sec:robustchallenges}, and then formulate the learning problem as continual learning and introduce our proposed method in \sref{sec:algorithm}. 
\subsection{Challenges for Learning Robust Tool Morphology}
\label{sec:robustchallenges}
We aim to learn a tool morphology that is customized for a specific contact-rich manipulation task, while also being robust and generalizable to task variations, e.g. different start states for the task. Instead of being able to handle just one initial state, as in~\cite{xu2021end}, we propose to learn a tool morphology parameter vector $\btheta$ that generalizes across a distribution of initial states $\bs_0 \sim \rho_0(\bs)$ . Since evaluating the expectation of the loss would be intractable,  we instead sample a set of $N$ initial states, $S = \{\bs_0^i| \bs_0^i \sim \rho_0(\bs), i=1 \ldots N\}$ and optimize the empirical expectation of the loss:
\subshrink
\begin{eqnarray}
\begin{aligned}
\label{eq:loss_empirical}
\btheta^* &= \argmin_\btheta \frac{1}{N} \textstyle\!\!\sum_{\bs_0^i \in S} \big[ \mathcal{L}(\pi; \bs_0^i, \btheta)\big], \\
&~~~S \!=\! \{\bs_0^i| \bs_0^i \sim \rho_0(\bs), i=1 \ldots N\}
\end{aligned}
\end{eqnarray}

\begin{figure}[t]
    \vspace{10px}
	\begin{center}
		\includegraphics[width=0.9\columnwidth]{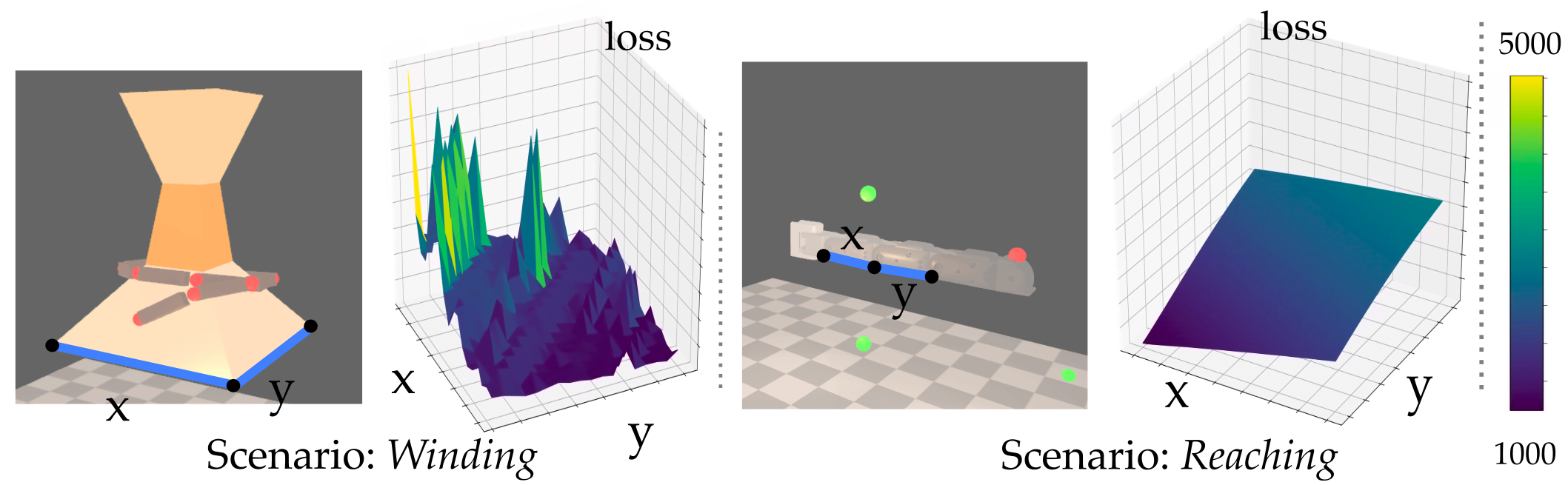}
		\caption{We visualize the 2D slices of the loss landscape for the contact-rich \textit{Winding} scenario and the no-contact \textit{Reaching} scenario. For \textit{Winding}, the goal is to prevent the rope from falling. For \textit{Reaching}, the goal is to optimize the arm so that the end effector can reach the green dots. The variables we optimize over are illustrated in the figure as x,y: for \textit{Winding} these are the lengths of two sides of the tool base; for \textit{Reaching} these are the lengths of two of the robot links. The details for the loss functions are given in the Appendix. It is clear that the optimization landscape of the contact-rich task is significantly more complex than that of the no-contact task.}
		\label{fig:landscape}
	\end{center}
\shrink
\vspace{-10px}
\end{figure}

Directly optimizing this loss with our pipeline in \sref{sec:pipeline} by unrolling episodes in the simulator with different initial states $\bs_0^i \in S$ presents several challenges. First of all, compared to tasks that do not have complex dynamics, contact-rich manipulation tasks usually have a highly non-convex optimization landscape, which causes gradient-based optimization algorithms to easily get stuck in local minima. \figref{fig:landscape} illustrates this by comparing the contact-rich task of \textit{Winding} a rope on a spool to a free-space \textit{Reaching} task. Therefore, for contact-rich tasks, directly optimizing \eref{eq:loss_empirical} might converge to local optima leading to suboptimal $\btheta$ values that determine the tool morphology. Second, there is a trade-off when choosing $N$, the number of samples in our empirical estimate of the loss in \eref{eq:loss_empirical}:
Small values of $N$ correspond to an insufficient number of samples, and would fail to capture the distribution $\rho_0(\bs)$. Large values of $N$ are computationally expensive, since evaluating even one gradient step requires unrolling $N$ episodes in the differentiable simulator. 

\subsection{Continual Learning Based Algorithm}
\label{sec:algorithm}
Given the challenges of learning a robust tool morphology with our end-to-end pipeline in \sref{sec:pipeline}, we draw inspiration from continual learning \cite{lesort2020continual, rebuffi2017icarl, mallya2018packnet}. 

Formally, we define a {\em scenario\/} as an MDP with an initial state distribution (see \sref{sec:pipeline}). Within a given scenario, we define a {\em task variation\/} $T^i$ as a restricted MDP with one initial state $\bs_0^i$ instead of a distribution over initial states. Therefore in this work, task variations are MDPs with different initial states. In general, task variations could be defined in other ways, e.g. variations of physical parameters or goal states. 
With sampling task variation $T^i$ using $\bs_0^i \sim \rho_0(\bs)$, optimizing \eref{eq:loss_empirical} is equivalent to minimizing the average loss on the sampled task set $T = \{T^1 \ldots T^N\}$: 
\subshrink
\begin{eqnarray}
\begin{aligned}
\label{eq:loss_empirical_tasks}
&\btheta^* 
= \argmin_\btheta \frac{1}{N} \sum_{T^i \in T} [\mathcal{L}(\pi; \bs_0^i, \btheta)], \\ &~T = \{T^1\ldots T^N\}
\subshrink
\end{aligned}
\end{eqnarray}
\prg{Morphology Optimization over a Sequence of Tasks} 
We can now formulate the problem of learning the morphology for a set of initial states as a continual, multi-task learning problem, where the morphology learned for one initial state can inform the optimization of morphology in other task variations with different initial states. 
To tackle the issue of this optimization being intractable for large values of $N$, we draw inspiration from continual learning and solve the problem for the sampled task set $T$ in a sequential manner. Task variations are solved sequentially in batches $B_0, B_1 ... $, where each batch $B_t$ contains $M$ task variations, $B_t = \{T_t^{(1)} \ldots T_t^{(M)} | T_t^{(i)} \in T, i = 1 \ldots M\}$. We select $M \ll N$ so that optimizing for each batch becomes tractable. By processing the batches sequentially, we address the trade-off outlined in \sref{sec:robustchallenges} between computational efficiency and sufficient coverage of task variations.

\prg{Constructing the Loss with Knowledge Distillation Regularization} We denote the morphology parameter vector we obtain after optimizing over a sequence of batches $B_1 ... B_{t-1}$ by $\btheta_{t-1}$. At timestep $t$, our method aims to learn $\btheta_t$ using an incremental learning strategy by processing the current batch $B_t$. Intuitively, we aim to minimize the task loss $L_t^{\text{task}}$ on the current batch $B_t$: 
    $L_t^{\text{task}}(\btheta) = \frac{1}{M} \sum_{T_t^{(i)} \in B_t} \mathcal{L}(\pi; T_t^{(i)}, \btheta).$
To avoid forgetting what has already been learned, we construct a regularization term $L_t^{\text{distill}}$ for distilling previous knowledge. To this end, we maintain a distillation task set $\mathcal{D}$, which is obtained by randomly sampling $M$ task variations from the previously seen ones, i.e. from $\bigcup \{ B_1, ..., B_{t-1} \}$.
When optimizing for $\btheta_t$ on the current batch, we still want the updated parameters $\btheta_t$ to perform similarly to $\btheta_{t-1}$ on the previous task variations in distillation set $\mathcal{D}$. Thus, we define the regularization term $L^{\text{distill}}$ as the squared error between the simulated trajectories generated when using $\btheta_t$ vs $\btheta_{t-1}$:
\begin{align*}
\label{eq:loss_batch_distill}
L_t^{\text{distill}}(\btheta_t) = \frac{1}{M} \!\! \sum_{T^{(i)} \! \in \mathcal{D}} &\Big(
sim(\pi; T_t^{(i)}, \btheta_{t-1}) \\
& - sim(\pi; T_t^{(i)}, \btheta_t) \Big)^2.
\end{align*}
We could compute the distillation loss on a part of the trajectory that matters for the task, e.g. height of the rope for the \textit{Winding} scenario (Appendix gives further details).

Finally, we get the overall loss by combining $L_t^{\text{task}}$ and $L_t^{\text{distill}}$ with a regularization coefficient $\alpha$: 
    $L^t(\btheta_t) = L_t^{\text{task}}(\btheta_t) + \alpha L_t^{\text{distill}}(\btheta_t).$
\vspace{7px}

\prg{Simplifying Optimization with Dimensionality Reduction} As visualized in \figref{fig:landscape}, the contact-rich tasks considered in this paper result in a complex optimization landscape that is difficult to optimize over. To alleviate this issue, we draw inspiration from Coordinate Descent \cite{wright2015coordinate} and propose to simplify the optimization problem by reducing the dimensionality of the decision variable. For a new batch $B_t$, we evaluate the gradient with respect to the currently optimal $\btheta_{t-1}$ and select $d^\prime < d$ dimensions of parameters $\btheta_{t-1}$ with the largest gradient magnitude. Then we only update these $d^\prime < d$ dimensions of parameters $\btheta_{t-1}$ to get $\btheta_t$. Once a dimension has been optimized in previous batches, we do not select it anymore. When all dimensions have been optimized, we restart (i.e. mark all dimensions as available for optimization) and repeat the process, decaying the learning rate with $\epsilon=e^{-1}$.
\section{Experiments}
\label{sec:experiments}
To test the performance of our proposed algorithm, we learn the tool morphology for three scenarios in simulation and evaluate the learned tool both in simulation and on a real robot manipulation setting. We first discuss the aspects that are the same for the three scenarios we consider, and then further elaborate on each of the scenarios separately.

\prg{Setup} We learn the tool morphology for all three scenarios with the differentiable simulator DiffHand \cite{xu2021end} \mx{and use L-BFGS-B as our optimizer implemented using Scipy \cite{virtanen2020scipy}}.
The implementation details including hyperparameters are given in the Appendix.

\prg{Scenarios} We consider three scenarios shown in  \figref{fig:sim_results}. 
\begin{enumerate}[leftmargin=0.5cm]
\item \textit{Winding}: In this scenario, a rope is wound around a tool. If the rope does not slip off the tool for a range of tool orientations, we consider this a success.  \mx{For \textit{Winding}, task variations correspond to the initial orientation of the rope and tool sampled from the uniform distribution over the space of 3D rotations i.e. $SO(3)$.} As a robot policy, we use a simple circular motion. We compute the loss after the rope is wound and the end of it is left hanging free. Unless supported by the tool, the rope will fall off. Since the simulator we use in this work can only model rigid objects, we use a chain of cuboids to approximate the rope in this task. 

\item \textit{Flipping}: In this scenario, we aim to learn a robot arm morphology that can flip a box by $90^{\circ}$. For control, here we first learn a basic flipping policy $\pi$ (described in Appendix). This scenario is similar to that in previous work~\cite{xu2021end}, except we consider a distribution of initial box poses. \mx{In prior work, the box is placed at one fixed position and fixed orientation facing the robot arm. Here, the cube position is randomly sampled from a square region. The cube orientation is sampled uniformly between $[-90^{\circ}, 90^{\circ}]$ around the z-axis.} 

\item \textit{Pushing}: This scenario is based on bimanual scooping used for food acquisition as in~\cite{grannen2022learning}. We aim to learn the morphology of a pusher that can push a pea on a table into a scoop. Peas are modelled as spheres. Task variations $T^i$ are instantiated with different initial pea positions \mx{ sampled uniformly from a square region}. For the trajectory of the pusher we use a forward motion with a zig-zag shape, which can make it challenging for the pusher to prevent the peas from rolling away. 
\end{enumerate}
We selected these scenarios because they include (\romannum{1}) various types of objects (rope, box, spherical peas), (\romannum{2}) objects with different physical properties (boxes with sharp edges in scenario \textit{Flipping}, smooth spheres in scenario \textit{Pushing}), and (\romannum{3}) varying contact conditions including a rope sliding on a winding tool, a robot holding on to a sharp box edge for pivoting, and spheres rolling over a tool's surface. 

\prg{Baselines}
We implemented two baselines for comparison: 
\begin{enumerate}[leftmargin=0.5cm]
    \item \textit{Baseline-DiffHand}: we optimize $\btheta$ for only one batch of tasks: $\btheta = \argmin  L_0^{\text{task}}(\btheta)$. 
    This baseline is a direct extension of DiffHand~\cite{xu2021end} obtained by replacing the single initial state with one batch of initial states and optimizing until convergence, without bringing in the continual learning aspect. 
    \item \textit{Simple-Continual}: we optimize $\btheta$ by minimizing the task loss sequentially for batches $B_1, ...,  B_t, ... $ as we would in a continual learning setting, making this a stronger baseline. At batch $B_t$, we start from $\btheta_{t-1}$ and obtain $\btheta_{t}$ using  $\btheta_t \!=\! \argmin  L_t^{\text{task}}(\btheta)$.   
    After optimizing for all the batches, we get the final morphology parameter $\btheta_{t=N/M}$. With the continual learning setting introduced, the difference between \textit{Ours} and this baseline is that \textit{Ours} uses knowledge distillation regularization in addition to the task loss, and simplifies the optimization with dimensionality reduction. 
\end{enumerate}

\subshrink
\subsection{Results \& Analysis in Simulation}
\label{sec:sim_results}
\begin{figure}[t]
    \vspace{10px}
	\begin{center}
		\includegraphics[width=\columnwidth]{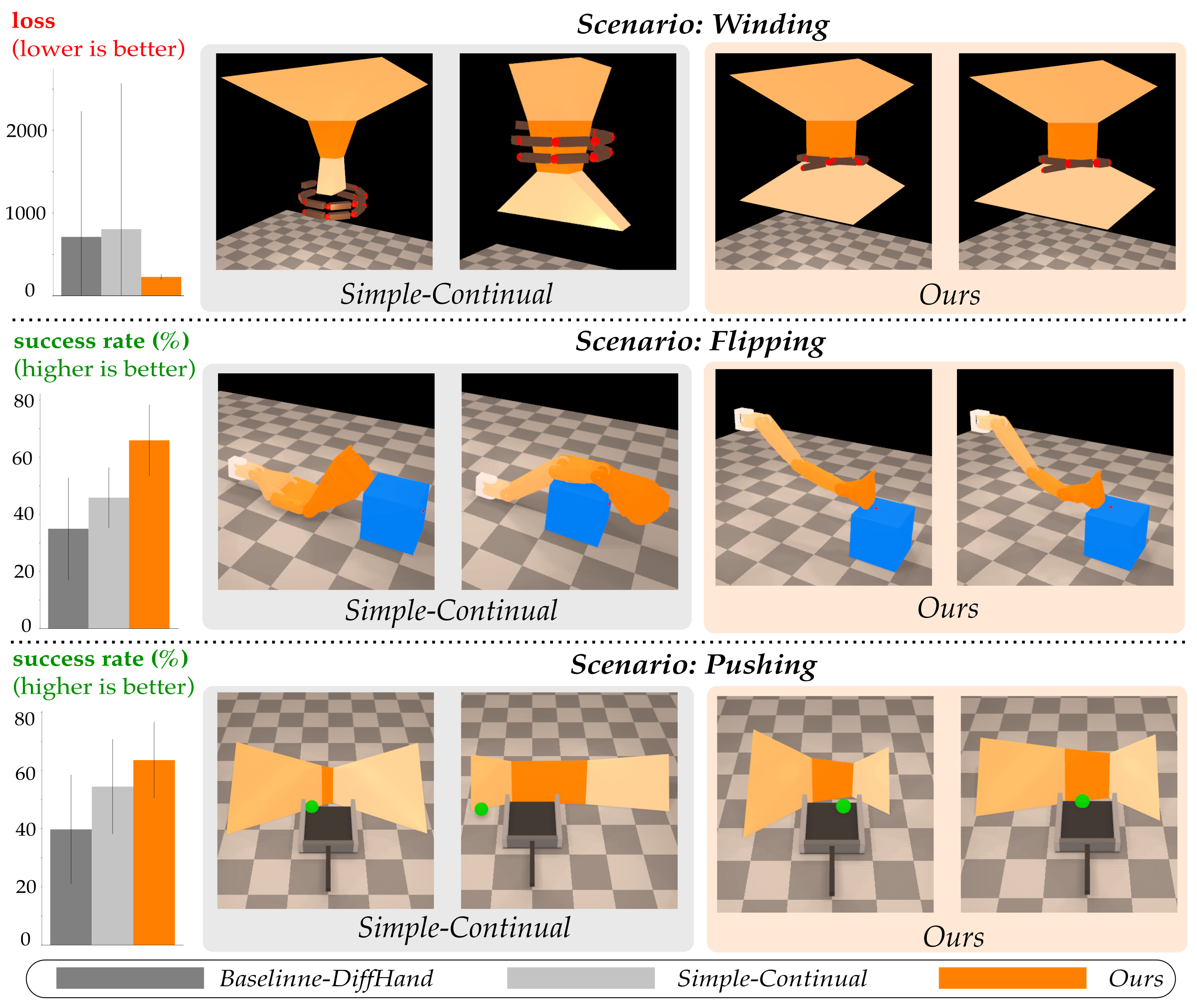}
		\caption{Evaluation results on scenarios \textit{Winding} (top row), \textit{Flipping} (middle row) and \textit{Pushing} (bottom row). For each scenario, we implement  \textit{Baseline-DiffHand},  \textit{Simple-Continual} and our algorithm (\textit{Ours}) in simulation. For quantitative evaluation, we report the mean and standard deviation over ten runs for the task loss for \textit{Winding} and test accuracy for \textit{Flipping} and \textit{Pushing}. For qualitative evaluation we visualize examples of two optimized morphologies from baseline \textit{Simple-Continual} and our algorithm (\textit{Ours}).}
		\label{fig:sim_results}
	\end{center}
\vspace{-15px}
\end{figure}

As shown in \figref{fig:sim_results}, the simulation results demonstrate that the tool morphology learned with our algorithm consistently outperforms both \textit{Baseline-DiffHand} and \textit{Simple-Continual} in terms of a lower test loss and higher test success rate. Across all scenarios, \textit{Baseline-DiffHand} demonstrates the worst performance with the largest variance. This is because, compared to optimizing over sequential batches of task variations, approximating the task distribution with only one batch will inherently be noisier. In scenario \textit{Flipping} and \textit{Pushing}, the baseline \textit{Simple-Continual} outperforms the \textit{Baseline-DiffHand}. This also indicates that learning from a sequence of task variations instead of one batch is beneficial. 
In scenario \textit{Winding}, baseline \textit{Simple-Continual} and \textit{Baseline-DiffHand} show poor performance with a high average loss and large standard deviation, while \textit{Ours} achieves a low loss and small variance (visualized as standard deviation in the bar plots).  For some of the runs, \textit{Simple-Continual} and \textit{Baseline-DiffHand} have an exceptionally high loss due to completely failing to hold the rope, so the rope quickly falls down. For both baselines, the performance deteriorates significantly for some batch sequences by converging to a suboptimal morphology parameter. In contrast, our method is not sensitive to how batches are sampled.
 
We visualize two of the learned tool morphologies for {\textit Ours} and baseline \textit{Simple-Continual} for each scenario in \figref{fig:sim_results}.  In scenario \textit{Winding}, one of the learned shapes from \textit{Simple-Continual} has one pointed end and one flat end, making the tool not effective for a certain range of orientations. This shows that the shape of the tool learned with baseline \textit{Simple-Continual} sometimes does not perform well across all task variations. In scenario \textit{Flipping}, in contrast to the wide arm tip learned by the \textit{Simple-Continual} baseline, our method learns a finer triangular tip, which can exert enough pressure close to the edge of the box to cause it to flip, thus achieving a higher success rate. In scenario \textit{Pushing}, our method learns a tool morphology with widening ends and a middle segment has a similar width as the scoop. The baseline \textit{Simple-Continual} either learns a narrow middle segment or a wide one. An appropriate width for the middle segment can better facilitate pushing the peas all the way into the scoop. Specifically, a too narrow middle segment might result in a gap between the pusher and the scoop and thus have a hard time pushing the peas all the way into the scoop, while a too wide middle segment can result in the pea rolling away.   
In summary, across the three scenarios, we observe that the learned morphology from our algorithm demonstrates better performance. 
 
\subshrink
\subsection{Evaluating Learned Tools in the Real World}
\begin{figure}[t]
    \vspace{5px}
	\begin{center}
		\includegraphics[width=0.95\columnwidth]{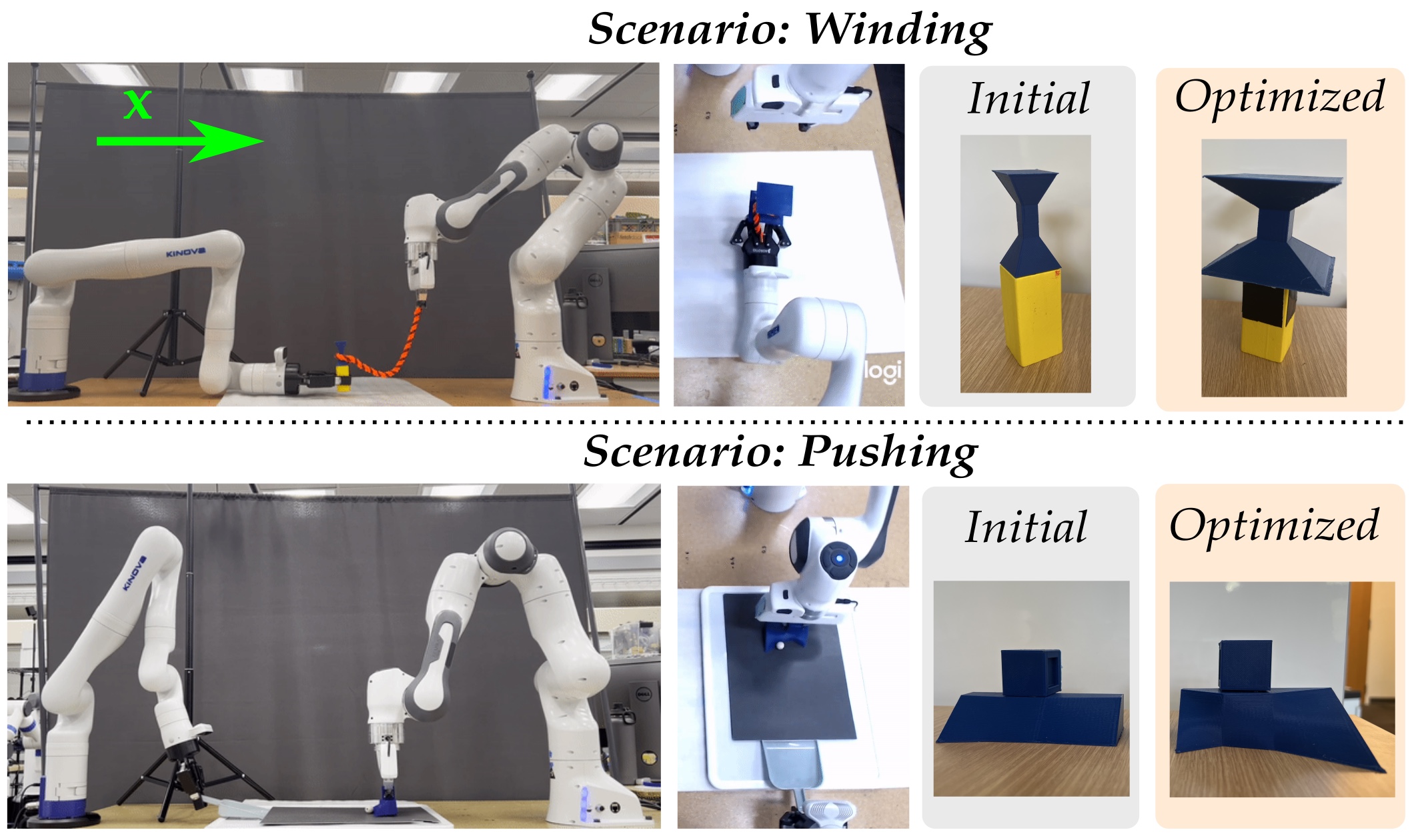}
		\caption{Real world experiment set up for \textit{Winding} and \textit{Pushing}. For each scenario, we 3D print the tool with the initial shape and the optimized shape obtained by our approach. For scenario \textit{Winding}, we first wind the rope around the tool and then rotate the tool around the highlighted x axis for 360 degrees to test whether the rope drops. For videos of experiments please see \href{https://sites.google.com/stanford.edu/learning-tool-morphology}{https://sites.google.com/stanford.edu/learning-tool-morphology}}
		\label{fig:real_setup}
	\end{center}
	\shrinkbottom
	\shrink
\end{figure}

We manufactured the tools learned with our method for scenario \textit{Winding} and \textit{Pushing} as shown in \figref{fig:real_setup}. The tools were 3D printed on an Ender 3 Pro 3D printer using PLA filament. We tested the functionality of the tool in these two scenarios. In scenario \textit{Winding}, we execute a simple control sequence that winds the rope around the tool. After winding, we rotate the tool in the air with a full circle of 360 degrees around the $x$ axis as shown in \figref{fig:real_setup} . A trial is counted as success if the rope successfully stays wrapped around the tool after winding and rotating. We run 5 episodes with the initial shape and the optimized shape. Results show that the optimized shape achieves a $100\%$ success rate while the initial shape keeps the rope on the tool only $20\%$ of the time. For scenario \textit{Pushing}, as shown in \figref{fig:real_setup}, we use the Franka Panda robot arm on the right to hold the pusher and try to push the white sphere to the flat dustpan held by a Kinova Gen3 arm on the left. The arm follows the waypoints of a pre-defined zig-zag trajectory. We also sample 4 initial locations for the pea and run 5 episodes for each initial location, which leads to 20 episodes for one tool. 
 The experiments show that the optimized shape achieves a $70\%$ success rate across the 20 trials while the initial shape achieves a $15\%$ success rate. 
 
\section{Conclusion}

To summarize, in this work, we approach the problem of custom tool design for robot manipulation in an end-to-end manner by leveraging the advantages of differentiable simulation. To learn versatile tool morphologies, we propose a continual learning approach that enables optimization over task variations, e.g. by varying initial object poses. We show that tools optimized with our method help to achieve an improved task performance compared to baselines in simulation. We also demonstrate that these tools enable successful task completion in reality.

\bibliography{references}  

\newpage
\section*{Appendix}

Here, we provide additional details for the \textit{Winding}, \textit{Flipping}, and \textit{Pushing} scenarios discussed in Section V and the \textit{Reaching} scenario visualized in Fig. 4. 
Recall that our objective is to optimize a vector of parameters $\btheta$, which encodes the morphology of the tool in each scenario. Across all scenarios in Section V, for our algorithm (\textit{Ours}), we set the regularization coefficient $\alpha\!=\!0.1$. We summarize other hyperparameters for each scenario in Table~\ref{tab:hyperparameters}. 
\vspace{5px}
\begin{table}[!h]
\begin{center}
\scriptsize
\begin{tabular}{ccccc}
\hline \noalign{\vskip 1mm}
Scenario     & $N$ & $M$ & $d$ & $d^\prime$   \\ \noalign{\vskip 1mm}\hline \hline\noalign{\vskip 1mm}
Winding   & 200  & 5  & 8  &  2                        \\ \noalign{\vskip 1mm}\hline\noalign{\vskip 1mm}
Flipping & 100  & 5  & 9  &  2                         \\ \noalign{\vskip 1mm}\hline\noalign{\vskip 1mm}
Pushing  & 100  & 5  & 7  &  2                       \\ \noalign{\vskip 1mm}\hline
\end{tabular}
\end{center}
\vspace{-10px}
\caption{Hyperparameters.} 
\label{tab:hyperparameters}
\end{table}

\subsection{Winding}
\begin{figure}[b]
    \vspace{10px}
	\begin{center}
		\includegraphics[width=\columnwidth]{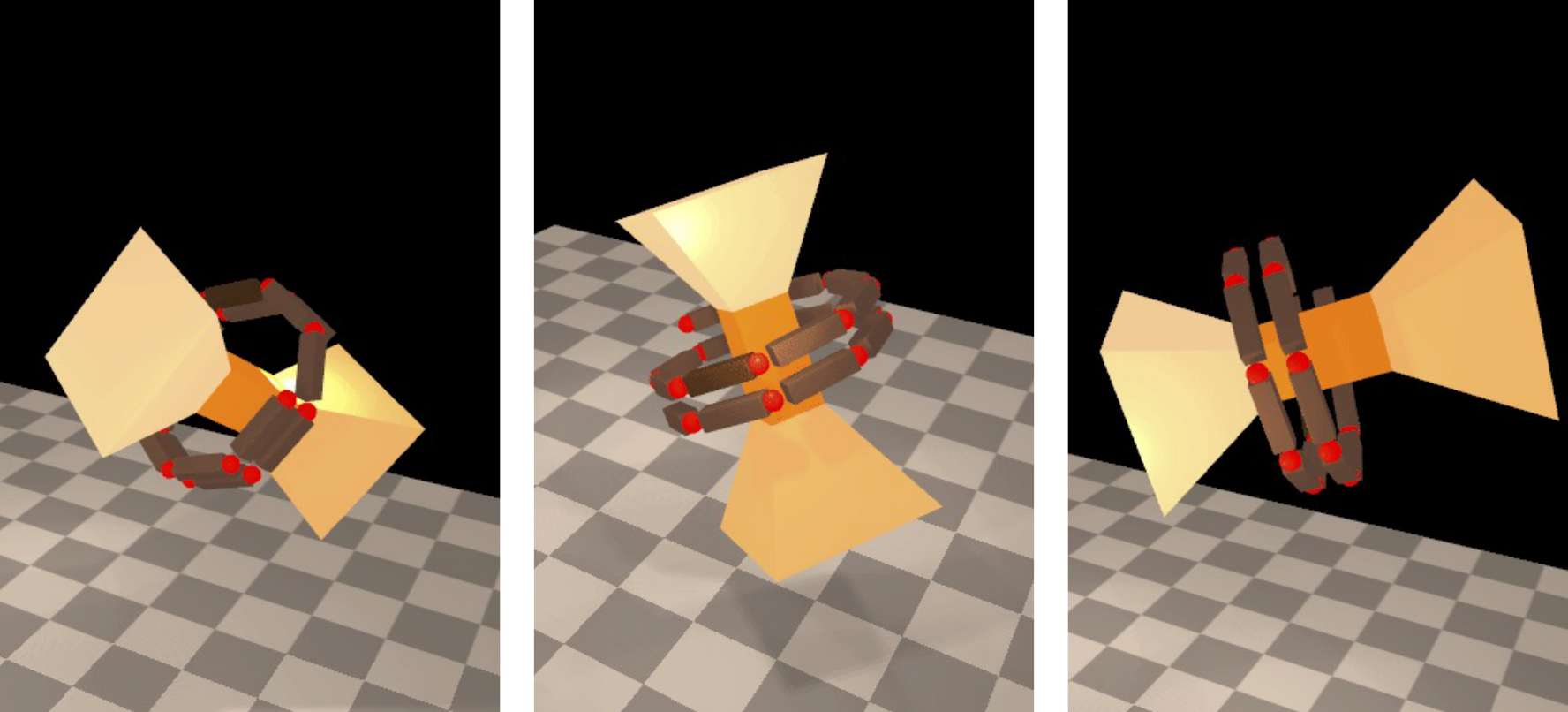}
		\caption{We visualize three example initial states for the scenario \textit{Winding}. Here, the initial orientation for the rope and the tool are sampled from a uniform distribution on the space of rotations in 3D.}
		\label{fig:rope_variations}
	\end{center}
\shrink
\end{figure}

\textbf{Loss function:} In this scenario, we use 15 linked cuboids to approximate a rope. At each simulation time step $\tau$, we denote the position of the rope's center of mass as $(x_{\tau}(\btheta), y_{\tau}(\btheta), h_{\tau}(\btheta))$. 
The task loss is the height of the rope's center of mass $h_{\tau}(\cdot)$ squared, summed over all time steps. This is computed by letting the rope fall under gravity for $H$ simulation steps:
\begin{equation*}
    L^{\text{task}}(\btheta) = \sum_{\tau=1}^H (h_{\tau}(\btheta) - h_0)^2.
    \label{eq:loss_task_winding}
\end{equation*}
Here, $h_0$ is the height of the rope's center of mass at time step $\tau=0$ at the start of the simulation. 
Suppose the  morphology optimized so far is expressed by $\btheta_{t-1}$. Our distillation loss for $\btheta_t$ to prevent our model from forgetting what has already been learned is defined as:
\begin{equation*}
    L^{\text{distill}}(\btheta_{\tau}) = \frac{1}{H} \sum_{\tau=1}^H (h_{\tau}(\btheta_k) - h_{\tau}(\btheta_{t-1}))^2.
    \label{eq:loss_task_winding}
\end{equation*}

\textbf{Policy:} In scenario \textit{Winding}, we initialize the rope to be placed around the tool. The task variations correspond to different initial orientation of the tool and rope as shown in \figref{fig:rope_variations}. We let the rope drop and the rope will not fall if the tool can effectively support the rope.

\subsection{Flipping}
\textbf{Loss function:}
\begin{figure}
    \vspace{10px}
	\begin{center}
		\includegraphics[width=\columnwidth]{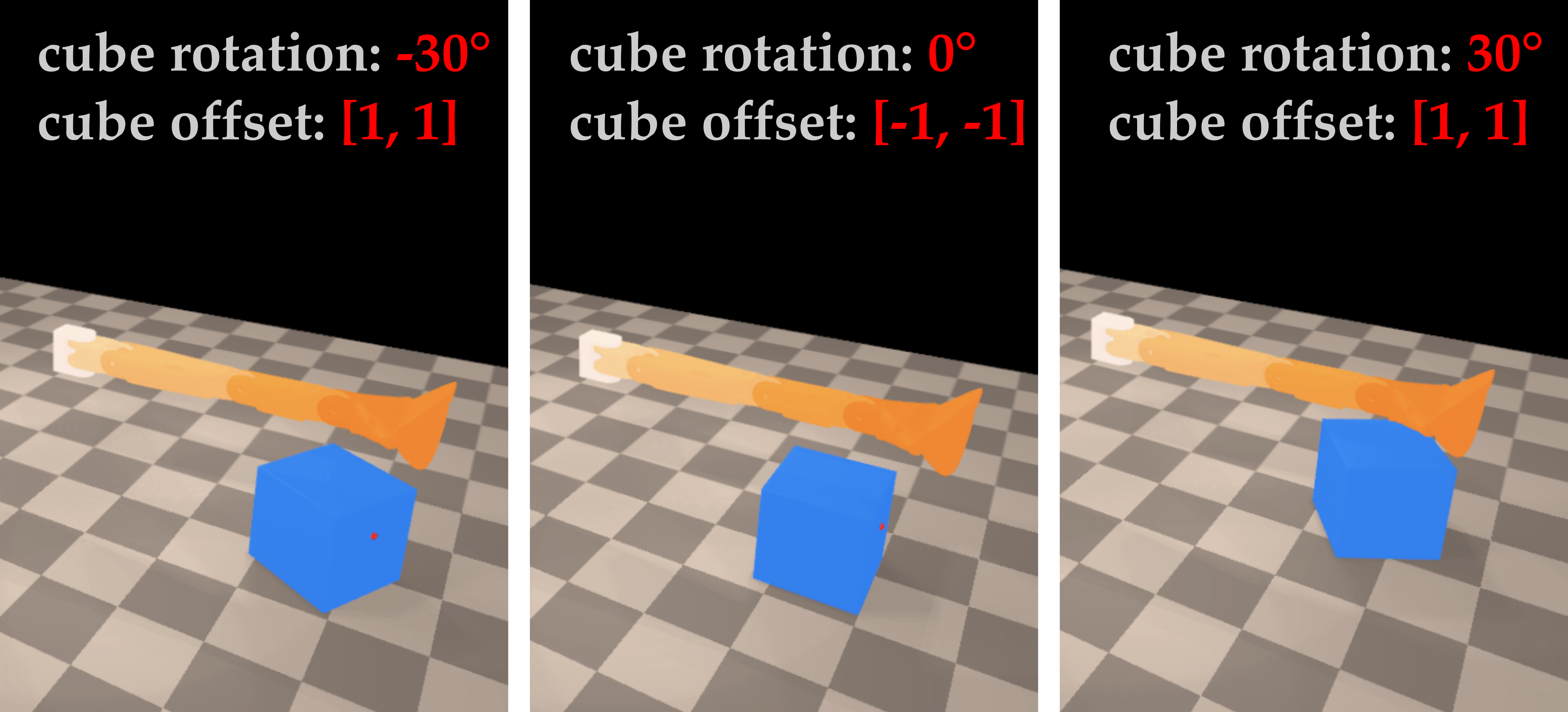}
		\caption{We visualize three example initial states for the scenario \textit{Flipping}. Here, the initial position is uniformly sampled from a square area in [-2, -2] to [2, 2], and the orientation for the box is sampled from a uniform distribution between $[-90^{\circ}, 90^{\circ}]$ around the vertical axis. }
		\label{fig:flip_variations}
	\end{center}
\shrink
\end{figure}
This scenario is adopted from~\cite{xu2021end} with the difference that we randomize the initial state of the box (see \figref{fig:flip_variations}). We use the same task loss as in~\cite{xu2021end} for $L^{\text{task}}$:
\begin{eqnarray*}
\begin{aligned}
L_{\text{task}}(\btheta)&=c_{\text {flip }}\left\|\phi_H(\btheta)-\frac{\pi}{2}\right\|^{2} \\
&~~~+\sum_{\tau=1}^H \left(c_{u}\left\|u_{\tau}(\btheta)\right\|^{2}+c_{\text {touch }}\left\|p_{\tau}(\btheta)-p_{\text {box }}\right\|^{2}\right) \\
&~~\text { with } c_{u}=5, c_{\text {touch }}=\left\{\begin{array}{ll}1 & t<H/ 2 \\ 0 & t \geq H / 2\end{array}, c_{\text{flip}}=50.\right.
\end{aligned}
\end{eqnarray*}
Here, for simulation step $\tau$, $u_{\tau}(\btheta) \in [-1,1]$ is the robot action, $p_{\tau}(\btheta)$ is the finger tip position, and $\phi_{\tau}(\btheta)$ is the rotation angle of the box.

For our algorithm, we define the loss $L^{\text{distill}}(\btheta_t)$ as:
\begin{eqnarray*}
\begin{aligned}
    L^{\text{distill}}(\btheta_t) &= \frac{1}{H} \sum_{\tau=1}^H \Big[ (u_{\tau}(\btheta_t) - u_{\tau}(\btheta_{t-1}))^2 
    \\&+ (p_{\tau}(\btheta_t) - p_{\tau}(\btheta_{t-1}))^2 +(\phi_{\tau}(\btheta_t) - \phi_{\tau}(\btheta_{t-1}))^2 \Big].
\end{aligned}
\end{eqnarray*}

\textbf{Policy:} For this task, we train a neural network (NN) to obtain a basic closed-loop policy $\pi$. For training data, we first randomly sample a range of starting cube orientations. Then for each pose, we separately run DiffHand~\cite{xu2021end} to learn a basic morphology and an open-loop policy. In practice, only a small number of such open-loop policies succeed. Hence, we did not attempt to jointly learn a policy and morphology over a distribution of starting poses, since that problem would be even more difficult. Nonetheless, we can use the trajectories that succeed at flipping the cube to construct a training dataset. With that, we train a basic NN policy that learns to imitate successful trajectories. This policy is `basic' for two reasons. First, only a small number of open-loop policies are successful (as mentioned above), so the training data contains example trajectories for only a small portion of the space. Second, the examples are from policies trained jointly with morphologies. This means the open-loop policies are unlikely to succeed when used with another morphology, unless we can solve the problem of finding a versatile morphology that works for a range of cube poses. The latter is exactly the problem we address in this work.
\vspace{5px}

\subsection{Pushing}
\begin{figure}[t]
	\begin{center}
		\includegraphics[width=0.6\columnwidth]{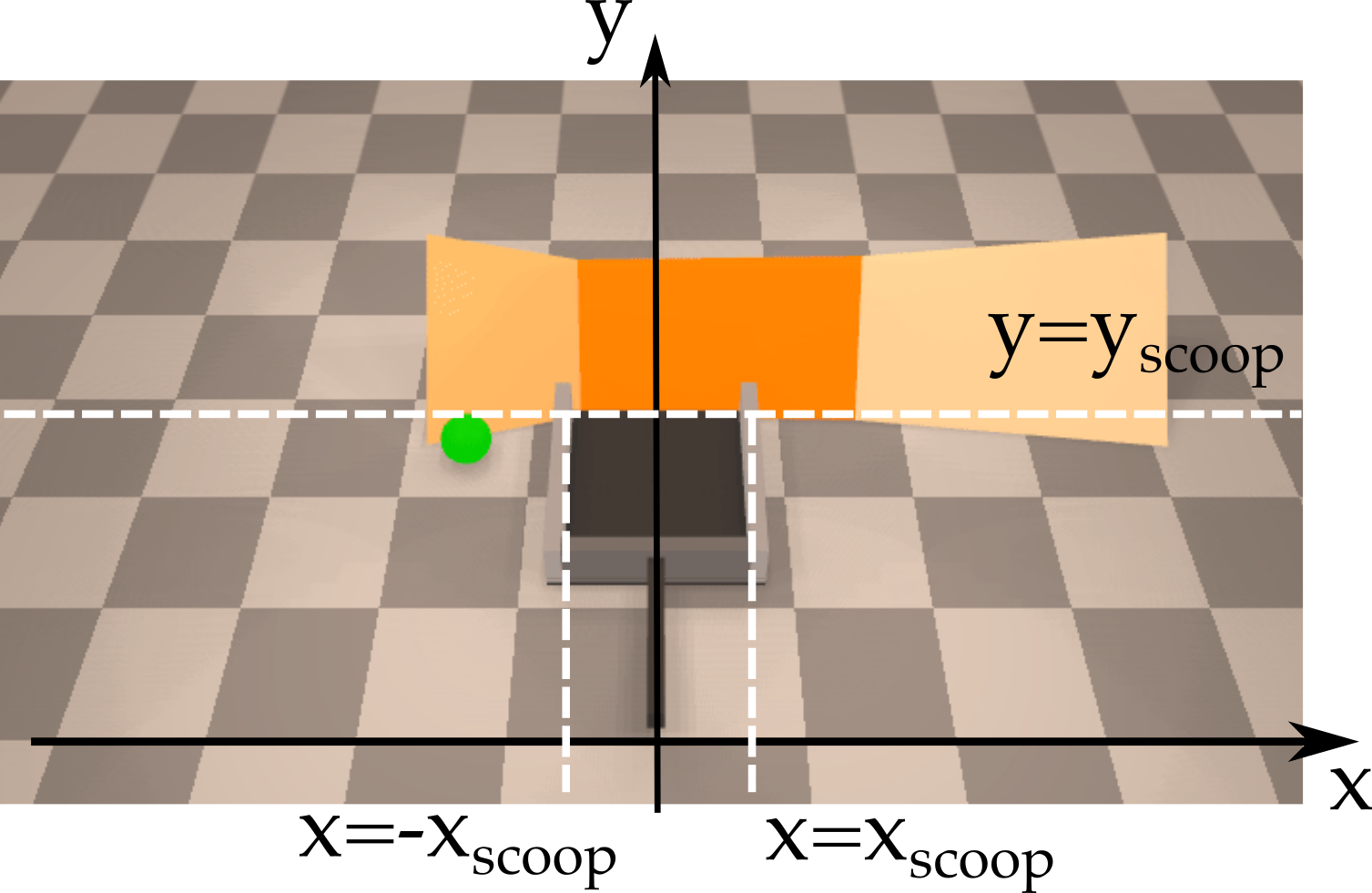}
		\caption{Visualization of \textit{Pushing} annotated with the scoop position. The tip of the opening of the grey scoop is at $\{(x,y) | y=y_{\text{scoop}},  -x_{\text{scoop}}<x<x_{\text{scoop}}\}$. Here, the green pea falls outside of the scoop, and thus will incur non-zero loss according to \eref{eq:loss_task_pushing}.}
		\label{fig:loss_pusher}
	\end{center}
 \vspace{-25px}
\end{figure}

\textbf{Loss function:} In this scenario, our goal is to push a pea onto a scoop that is placed on the table. We denote the half width of the scoop to be $x_{\text{scoop}}$. \figref{fig:loss_pusher} visualizes the scoop placed such that its opening is located at $\{(x,y) | y=y_{\text{scoop}},  -x_{\text{scoop}}<x<x_{\text{scoop}}\}$. We denote the x coordinate of the pea's position as $\Tilde{x}(\btheta)$. In the figure the y coordinate of the pea's position is $y=y_{\text{scoop}}$, same as the y coordinate of the tip of the scoop.  We construct the task loss by giving a penalty when the pea is outside of the opening of the scoop, i.e. $\Tilde{x}_i(\btheta) \not \in (-x_{\text{scoop}}, x_{\text{scoop}})$:

\begin{equation*}
    L^{\text{task}}(\btheta) =     \left \{\begin{array}{ll} 0 & ||\Tilde{x}(\btheta)|| < x_{\text{scoop}} \\  (||\Tilde{x}(\btheta)|| - x_{\text{scoop}})^2 & ||\Tilde{x}(\btheta)|| \geq x_{\text{scoop}} \end{array}  \}\right.
\label{eq:loss_task_pushing}
\end{equation*}

We define the distillation loss as:
\begin{equation*}
    L^{\text{distill}}(\btheta_t) = \sum_{\tau=1}^H (x_{\tau}(\btheta_t) - x_{\tau}(\btheta_{t-1}))^2 + (y_{\tau}(\btheta_t) - y_{\tau}(\btheta_{t-1}))^2.
\end{equation*}

\textbf{Policy:} For this scenario, we use a predefined zig-zag trajectory for the pusher. This motion makes it more challenging for the pusher to prevent the peas from rolling away.

\subsection{Reaching}
We take this scenario from \cite{xu2021end} for visualizing the landscape in Fig. 4. Here, a finger with multiple joints is assumed to be mounted on a wall. The finger aims to sequentially reach the target points represented by the green dots in Fig. 4. For simulation step $\tau$, $u_{\tau}(\btheta)  \in [-1,1]$ is the action, $p_{\tau}(\btheta)$ is the finger tip position and $\hat{p}_{\tau}$ is the target point. The task loss is computed by:
\begin{equation*}
    \begin{array}{r}\mathcal{L}^{\text{task}}(\btheta)=\sum_{\tau=1}^{H} c_{u}\left\|u_{t}(\btheta)\right\|^{2}+c_{p}\left\|p_{t}(\btheta)-\hat{p}_{t}\right\| \\ \text { with } c_{u}=0.1, c_{p}=10.\end{array}
\end{equation*}

\end{document}